\documentclass{article}

\PassOptionsToPackage{square, numbers, compress}{natbib}

\usepackage{graphicx}
\usepackage{epsfig}
\usepackage{dsfont}
\usepackage{bbm}
\usepackage{booktabs}
\usepackage{multirow}

\usepackage[final]{neurips_2022}

\usepackage[utf8]{inputenc} 
\usepackage[T1]{fontenc}    
\usepackage{hyperref}       
\usepackage{url}            
\usepackage{booktabs}       
\usepackage{amsfonts}       
\usepackage{nicefrac}       
\usepackage{microtype}      
\usepackage{xcolor}         
\usepackage{commath}
\usepackage{amsmath,amssymb}
\usepackage{algorithm}
\usepackage[noend]{algpseudocode}
\usepackage{mathtools}
\usepackage{tabularx}
\usepackage[caption=false]{subfig}
\title{Cut-and-Approximate: 3D Shape Reconstruction from Planar Cross-sections with Deep Reinforcement Learning}

\author{%
  Azimkhon A.~Ostonov\\
  Department of Computer, Electrical and Mathematical Science \& Engineering\\
  King Abdullah University of Science and Technology\\
  Thuwal, SA 23955-6900\\
  \texttt{azimkhon.ostonov@kaust.edu.sa} \\
}

\begin{document}

\maketitle

\begin{abstract}
Current methods for 3D object reconstruction from a set of planar cross-sections still struggle to capture detailed topology or require a considerable number of cross-sections. In this paper, we present, to the best of our knowledge the first 3D shape reconstruction network to solve this task which additionally uses orthographic projections of the shape. Our method is based on applying a Reinforcement Learning algorithm to learn how to effectively parse the shape using a trial-and-error scheme relying on scalar rewards. This method cuts a part of a 3D shape in each step which is then approximated as a polygon mesh. The agent aims to maximize the reward that depends on the accuracy of surface reconstruction for the approximated parts. We also consider pre-training of the network for faster learning using demonstrations generated by a heuristic approach. Experiments show that our training algorithm which benefits from both imitation learning and also self exploration, learns efficient policies faster, which results the agent to produce visually compelling results.
\end{abstract}

\section{Introduction}

Surface reconstruction from planar cross-sections has many applications including organ reconstruction from signals stemming from MRI scanners, CT scanners \cite{9434040}, terrain contour lines \cite{10.1016/j.cad.2008.01.005}, orebody modeling, and many others. The problem in these examples can be formulated as building a manifold surface interpolating (or approximating) given planar cross-sections. More concretely, the task is equivalent to classifying  each point $p \in R^{3}$ as outside, inside or on the surface taking into account the given input information, so that classification results should match for each individual cross-section.\\
In many cases, several input planar cross-sections do not provide sufficient information about the geometry and topology of the object. This makes it hard to detect connectivity and branching cases between two consecutive cross sections. Increasing the number of cross sections is not an optimal solution as it could result in slow performance of the reconstruction algorithm or additional data might simply not be available. However, in many surface reconstruction domains, there exists some prior information representing detailed geometry and topology of the objects that could be utilized. This prior information in our case consists of $3$ different orthographic projections (front view, top view and end view) of the object. One particular advantage of this representation is that in most cases it can be drawn from the initial slices, which makes the given data sufficient without having additional knowledge.\\
Orthographic projections of the shape represent enough valuable information about its topology and geometry. Our approach is next to employ classical divide-and-conquer algorithm to effectively parse the initial shape into smaller regions (bounded by primitives) and then reconstruct each part separately. Shape parsing as a collection of smaller parts is an important aspect of shape understanding and analysis \cite{Gopal2017, abstractionTulsiani17,Cheng2020,Despiona2019}. This is a first step for computer programs or robots to understand 3D shape. The result of the parsing of the input shape into smaller parts then can be used in many applications to further analyze important features for example, stability of the input shape, geometrical features such as symmetry and connectivity. More detailed shape parsing therefore constitutes as a foundation for more accurate estimation of these features. As a result of this step, in the ideal case, we assume that each primitive (box in our case) bounds a part of the object which has no branches and disconnections. For this type of object most surface reconstruction algorithms work efficiently \cite{boissonnat1988shape, bajaj1996arbitrary}. Then, the final result would be possible to get by combining all reconstructed parts.\\

\begin{figure}
  \centering
  \includegraphics[width=1.0\columnwidth]{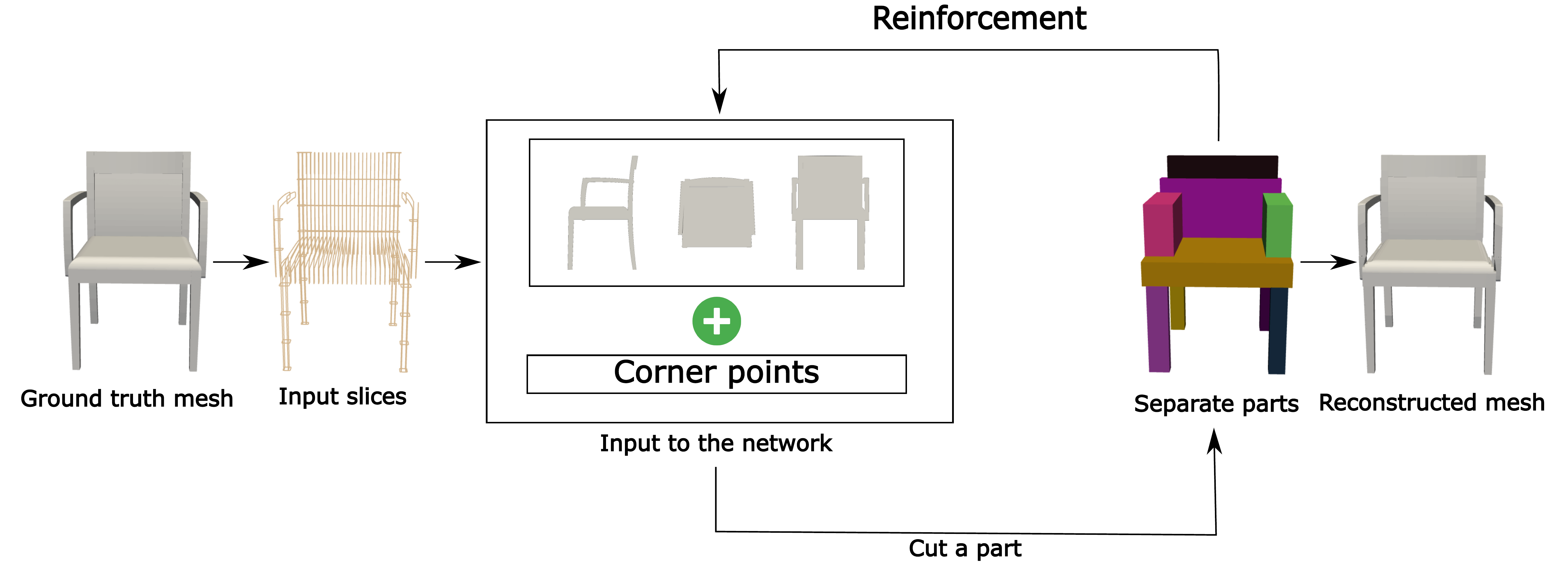}
  \caption{The pipeline for shape reconstruction from slices. We assume that the initial slices and orthographic projections of the shape into 3 planes are given. Our method based on Reinforcement Learning iteratively divides the initial shape into smaller parts to decrease reconstruction error. Then separate parts are reconstructed and combined to obtain the final result.}
  \label{fig:pipeline}
\end{figure}

Our approach for the shape parsing is neural network based and it works sequentially, cutting a part of the shape until all the area is covered. Taking into account input data representation we consider using corner points detection algorithm in the orthographic projection images to help the agent. Corner points usually indicates to complex geometry or branching points in the object topology. We learn effective shape parsing which leads to increase in the accuracy of the reconstruction from planar cross-sections of the shape. We employ Reinforcement Learning (RL) to solve this task, by taking actions and collecting rewards. In each step we cut a part of 3D object and reconstruct it from planar cross-sections in this region. The reward function is designed to take into account the accuracy of the reconstruction. Objects are arbitrarily selected from the ShapeNet \cite{Chang2015} dataset and the number of planar cross sections is fixed. In the experiments section we show that our algorithm works efficiently for both ShapeNet objects and also downloaded objects from the Internet.\\
One of the problems related with using this approach is computational complexity while working with 3D objects. Direct applying off-the-shelf RL methods does not produce satisfying results. Therefore, following the works \cite{VecerikHSWPPHRL17, Nair2017} we also consider to use Imitation Learning for quick-start pre-training of the network. The difference of our approach is we use a heuristic method to generate demonstration data and effectively search for optimal policies. This boosts the performance of our method and results in visually acceptable 3D reconstructions.\\
\indent Our contributions are as follows:
\begin{itemize}
    \item We propose topology-aware reconstruction of 3D models represented as planar cross-sections using an RL algorithm. To the best of our knowledge this is the first attempt using RL to sequentially cut parts of a 3D model to reconstruct a given shape.
    \item We propose an efficient training algorithm which benefits from both Imitation Learning (IL) and RL. This is mainly due to that we considered to update suboptimal demonstration data in the replay buffer using both agent's self exploration data and heuristic policy during the training.
\end{itemize}
We demonstrate the potential of our algorithm on the ShapeNet \cite{Chang2015} data set on various types of shape classes in terms of how it captures the details of the 3D model by comparing with state-of-the-art methods. Additionally, we access our method on different models downloaded from the Internet to show the generalization capability of the method.

\section{Related Work}
\textbf{Slice based input shape reconstruction.} 3D shape reconstruction from planar contours is well explored problem. This is especially popular in the medical image reconstruction. Most of these work focused on creating a closed surface from existing sections usually provided by CT or MRI radiologists. Earlier approaches are directed to solve the connectivity problem between parallel slices with different constraints on shape geometries \cite{boissonnat1988shape, wang1986surface}. Bajaj et al. \cite{bajaj1996arbitrary} considered arbitrary topology shape reconstruction from parallel planar cross sections. Liu et al. \cite{liu2008surface} proposed to use medial axes based space partitioning using given slices to reconstruct the shape from non-parallel contours. The main advantage of these methods is fast reconstruction, but they rely on a large number of slices and also the final result usually contains jagged areas which requires additional post-processing using mesh smoothing algorithms. Additionally, the above methods fail to find correspondence between consecutive slices with difficult geometries, branches, self intersections. Another common approach which studied in many previous work is related with extraction of the surface as the zero set of an implicit function \cite{turk2005shape, csebfalvi2002smooth, Bermano2011, Zou2015}. For example, Bermano et al. \cite{Bermano2011} proposed an effective approach to capture unknown regions based on the given data. However, problems related with branching and connectivity between difficult geometries still remain open. Few other approaches are based on simulated annealing \cite{pellot19943d, Ning2001}, which assumes infinite-source of slices from one or two orthogonal views are given. Each slice is then modelled as a 2D Markov-Gibbs random field which are concatenated to form 3D shape. These work also focused on the reconstruction of medical images. Zou et al. \cite{Zou2015} considered topology-constrained surface reconstruction from cross sections. One example of reconstructing a 3D  shape from slices is proposed by Fang et al. \cite{fang2020}, which uses point-cloud input. Shen et al. \cite{Shen_2019_CVPR} considered reconstruction of input 3D shape, predicting slices from image in the frequency domain. The output for their method is a volumetric mesh. Our work is different from these methods, we solve this problem in RL paradigm by obtaining rewards for each action.

\textbf{Using additional data for surface reconstruction.} The main problem related with surface reconstruction from planar cross sections are due to not awareness of the topological information of the shape, such as connectivity, branching and self-intersections. This type of information is hard to convey with slices as it would require large number of them. Thus, some previous methods \cite{liu2008surface, attene2013polygon, Zou2015, huang2017topology} rely on reconstructing the surface and then apply mesh fairing, smoothing algorithms to reduce the topological errors. However, this usually results overly smooth regions which is not compelling with the initial information characterized by the slices. Some other work considered to additionally utilize topological information. Most notably, Zou et al. \cite{Zou2015} proposed to use the number of genus in the shape, and then apply dynamic programming approach to find the most suitable topology out of several options based on handcrafted score function. While this approach solves some topological issues, connectivity and branching related problems still remains unsolved. Some methods addressed connectivity problems for closed surfaces \cite{nooruddin2003simplification, ju2007editing}. Others put this problem to solve by user interactions \cite{sharf2006competing, yin2014morfit}. One common approach is related with using templates to fit the resulting mesh to the ground-truth mesh \cite{holloway2016template}. Templates usually include different views, orthographic projections of the shape. In this work we use orthographic projections, as in many cases it can be derived from the slices, and also contain topological information about the shape.

\textbf{Shape generation and reconstruction with Reinforcement Learning.} RL is successively used in many 2D domains including stroke-based painting \cite{Ganin2018,Zhewei,John2019}, shape grammar parsing \cite{Teboul}, sketch drawing \cite{Tao2018,Umar2018}, scene synthesis \cite{ostonov2022rlss}, point-cloud registration \cite{inproceedingsBauer}. In 3D case, Sharma et al. \cite{Gopal2017} proposed a network based on an encoder-decoder architecture and RL to produce a compact program to generate 2D and 3D input shape. The input shape in this work is obtained from primitive shapes by applying Boolean operations formed as a grammar. However, the 3D shapes in the real life could be much more complex to obtain them with such a grammar, which limits the use of this method to simple shapes. While, we train our method on ShapeNet \cite{Chang2015} data set.
The work proposed by Lin et al. \cite{Cheng2020}, uses two networks to model 3D shape like human modelers. The first network Prim-Agent represents 3D shape as a combination of cuboids and the second network Mesh-Agent is used to smooth the output of the first network to obtain similar result to the input shape. This method also uses RL with imitation learning from expert demonstrations generated by heuristic approach using modified DAgger \cite{Ross2010} algorithm. Our work is different from their approach in many ways: 1) Our approach is based on more detailed reconstruction of the input shape using slices 2) The input to train the networks also not the same 3) We use continuous action space using DDPG and pre-training, while Lin et al. uses DQN \cite{DBLP:journals/corr/MnihKSGAWR13} which uses discrete action space.

\section{Method}
In this section we first consider corner points detection, the main aspects of Reinforcement Learning method, following virtual expert policy to accelerate training. Then, we finally shift to network training using both IL and RL for more accurate shape reconstruction.
\subsection{Corner Points Detection}
To reconstruct 3D shape from the planar cross-sections (or slices) $\{C_k\}_{k=1}^{n_{C}}$, we utilize its up to three orthographic projections: $I_{i}, i \in \{1, 2, 3\}$. In each step algorithm chooses one of the projections and selects a region in it to separate from the initial shape. At the end 3D shape is simply union of all the individual parts, $P = \bigcup\limits_{i=1}^{K} S_{i}$, $K$ is number of all parts. Each part to be selected approximated using slices in this region. In the optimal case, each the individual part consists of only one connected component without any branching. To achieve this, the first step is to detect the corner points in the projection images. Detecting corner points in the image is well explored topic \cite{harris1988combined,shi1994good,trajkovic1998fast,smith1997susan,mokhtarian1998robust,lowe2004distinctive,rosten2006machine}. The corner points detected using corner response function ($C$) for all pixels in the image region. This is computed by calculating the following:
\begin{align*}
M = \sum_{x,y} w(x,y)\begin{bmatrix}
I_{x}^2 & I_{x} I_{y}\\
I_{x} I_{y} & I_{y}^2
\end{bmatrix}\\ C = |M| - k(trace\ M)^2
\end{align*}
Where $w(x,y)$ is a window function, $k$ is a parameter, $I_{x}$ is the image derivative in the $x$ direction. When $C$ is larger than the given threshold (this happens when both eigenvalues of $C$, $\lambda_{1}, \lambda_{2}$ are large) then the corresponding pixel considered a corner point.

\subsection{Reinforcement Learning Main Aspects}
\begin{figure}
  \centering
  \includegraphics[width=1.0\columnwidth]{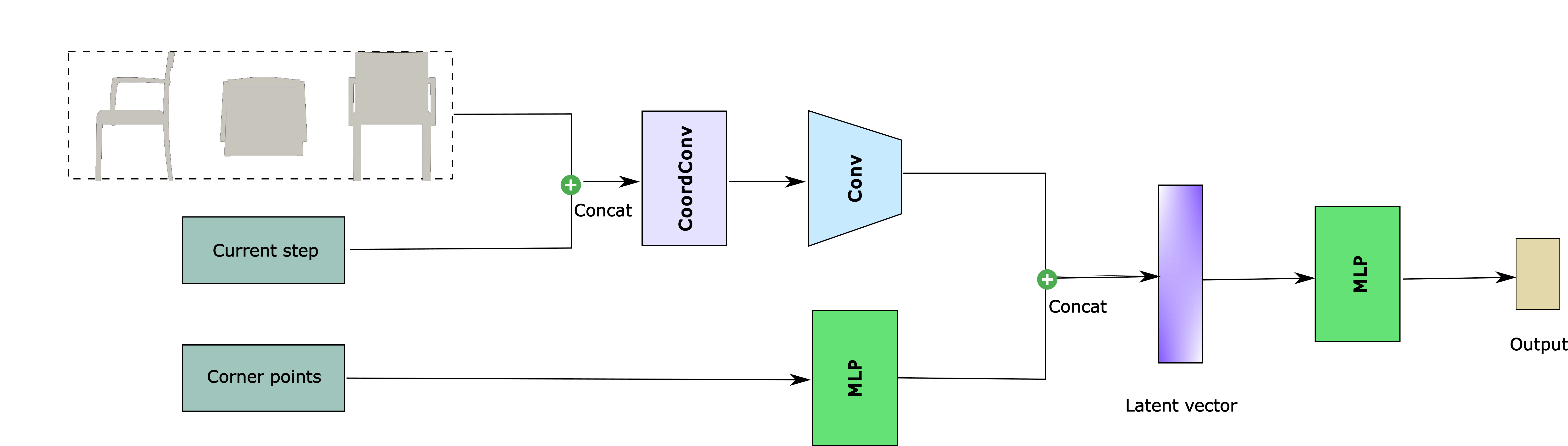}
  \caption{Network architecture of our method. The input is consists of 3 orthographic projections of the shape, the current step and the corner points in the images. We use \cite{10.5555/3327546.3327630} CoordConv before convolutional operations as the first layer. MLP here specifies fully connected networks, Conv refers to convolutional networks.}
  \label{fig:net_arch}
\end{figure}

The main part of our approach is based on sequentially cutting a region of 3D shape at each step and thereby approximating given shape by a combination of smaller shapes (parts) each represented as polygon meshes. The goal of the agent is to learn more accurate representation of the 3D shape from given scalar feedback.\\
\textbf{State}\ representation include 3 projections on the planes parallel to its sides, current step and also concatenated list of corner points corresponding to each of the projections Figure \ref{fig:net_arch}.\\
\textbf{Action}\ is defined by five-tuple $(x_1, y_1, x_2, y_2, c)$ which specifies coordinates of two points $P_1 = (x_1, y_1)$, $P_2 = (x_2, y_2)$, 
and number of a plane $c$. 
All of the fields of the tuple are continuous, and take values from $[0, 1]$ interval, these are then adjusted to the necessary scale. 
For each of these points we find closest corner point in the plane $c$, then we divide the initial shape by two, cutting along the line segment whose ends defined by these points. The smallest part out of these two sections we define with $W$. At the $i-$ th step $W_i$ is separated from the main part and reconstructed using slices in this region. The reconstructed part is defined with $S_{i}$. For the simplicity we only consider cuboids as the boundary to cut from the shape in this paper. The process is repeated until the initial shape $O$ is fully represented as summation of polygon meshes $\cup S_{i}$.\\
\textbf{Reward Function}\ takes into account consistency of the polygon mesh representation and initial shape $O$. Also, it encourages to cover bigger part of the initial shape at each step by rewarding parsimony. For the $i$-th step, it is defined as follows:
\begin{align*}
R_i = IoU(W_i, O) + \lambda \times IoU(S_i, W_i)
\end{align*}
$IoU(A, B)$ specifies intersection-over-union between two parts $A$ and $B$.

\subsection{Expert Policy}
Complex state representation which includes projections of 3D data in our case results to slow learning. This is even more complicated if we consider heavy intersection and union operations on 3D data, and mesh generation which needs to be repeated in each step. As a result we seek for solutions which do not require lots of iterations to learn shape reconstruction. One way out of this problem which we consider here is to incorporate Imitation Learning (IL) to start with accurate initial policy. In our case there is no expert generated data available. Therefore, we adopt heuristic policy which can serve as a quick-start for the training. First, we pair each corner point in $I_{i}, i \in \{1, 2, 3\}$ projection with its nearest neighbour and form line segments with endings at these points. Then, from these pairs select those which result in increase to the number of connected components when the corresponding line is removed. For each of the connected components we find bounding-box with minimal area. $A_{i}, i\in[1, K]$ ($K - $ is the number of the connected components) specifies the area of the part which lies in $i$th box and at the same time does not belong to $i$th connected component. Finally, sort all of the line segments $L_{i}, i\in[1, Q]$ belonging to all $3$ views by increasing order of corresponding $A_{i}$. The heuristic policy is to use this sorted line segment coordinates and appropriate projection as actions.

\subsection{Network training}
Heuristic search given above does not find the optimal solution, instead it can be used to pre-train the network to speed up learning. Our training mechanism benefits from both Imitation Learning and Reinforcement Learning.\\
\indent \textbf{Agent Training Method:} Taking into account continuous action space of the agent we adopt the Deep Deterministic Policy Gradient (DDPG) \cite{LillicrapHPHETS15} to implement the training. DDPG maintains both actor $\pi(s)$ and critic $Q(s, a)$ networks, with parameters $\theta_{\pi}$ and $\theta _{Q}$ respectively. Policy $\pi$ is a mapping between a state $s_t$ and action $a_t$. $Q(s_t, a_t)$ is action-value function, it estimates expected reward after taking an action $a_t$ in state $s_t$ and then following greedy policy afterwards. It can be computed from recursive relationship known as Bellman equation: 

$$Q(s_t, a_t) = r(s_t, a_t) + \gamma Q(s_{t+1}, \pi(s_{t+1}))$$

$r(s_t, a_t)$ specifies reward after taking action $a_t$ at the state $s_t$, $\gamma$ is discounting factor $\gamma \in [0, 1]$.\\
The actor network is updated to produce actions which maximize expected returns. Therefore, loss in this case is simply mean of $Q$-values for states. The critic network is updated with simple TD loss. As in DQN \cite{mnih2013atari,mnih2015humanlevel} DDPG also uses replay buffer and target network (in this case for both actor and ctitic) to stabilize learning.\\

\indent \textbf{Training Algorithm:} 
\indent Agent training algorithm in our case consists of two phases: I) Imitation Learning using heuristic policy II) learning from self exploration.\\
For the network training at the first phase we use heuristic algorithm described above. In this case it specifies policy $\pi^{h}$ and generates demonstrations for training. The difference of these demonstrations from \cite{xiaoqinpretrain2018} is it includes some additional parameters such as, reward data and coordinates of the second point ${(s_t, a_t, r_t, x_2, y_2)}_{t=0,1,2,...}$. Our pre-training algorithm is based on DDPGfD \cite{VecerikHSWPPHRL17}. In their work they consider solving robotic problems with sparse rewards using pre-training with expert demonstrations. There are two key differences between our work and \cite{VecerikHSWPPHRL17}: First, in our case we get dense rewards for our actions, Hence, we omit some losses in the objective function. Second, we use heuristic algorithm to generate demonstrations which might not be always optimal solution for the problem. Therefore, additionally we add Behavioral Cloning (BC) loss for training the actor with Q-filter \cite{Nair2017}. The difference here is we use heuristic policy generator. This is to eliminate non-optimal learning trajectories. 

$$L_{BC} = \sum_{i=1}^{n_{D}} \norm{\pi(s_i|\theta_{\pi}) - \pi^{h}(s_i)}^2 \mathds{1}_{Q(s_i , \pi^{h}(s_i) )> Q(s_i , \pi (s_{i}))}$$

$n_{D}$ refers to the minibatch size (it is equal to 64 in our experiments).\\
Use of BC loss requires availability of expert feedback during all of the time of training. This condition is applicable for our case since we use heuristic algorithm for data generation. As heuristic policy generator always available, we can also apply BC loss to the demonstration data and as well as agent's self exploration data. The final loss defined as a sum of losses for each actor and critic.

\indent \textbf{Using two replay buffers:} We keep two replay buffers $R_D$ for demonstration data and $R_A$ for agent's self exploration data. During the training process in each step two minibatches sampled from each of the replay buffers for the network training. Demonstration data in our work is generated by heuristic algorithm and is different from the robotic control tasks \cite{VecerikHSWPPHRL17, Nair2017} where demonstration data is gathered using real robots and human implication. Therefore, in robotic learning tasks it is meaningful to keep agent stick on demonstration data to avoid unrealistic scenarios. In this work, we use only efficient demonstration data.
Agent training process is given in the following Algorithm.1.

\begin{algorithm}
	\caption{Training Algorithm} 
    Randomly initialize critic $Q(s, a|\theta^Q)$ and actor $\pi (s|\theta^\pi)$ networks\\
    Initialize target network $Q^{'}$ and $\pi^{'}$ with weights $\theta^{Q^{'}} \gets \theta^{Q}$, $\theta^{\pi^{'}} \gets \theta^{\pi}$\\
    Initialize replay buffer $R_D$ for demonstration data and $R_A$ for agent's self exploration data
	\begin{algorithmic}[1]
	    \For {$t = 1, K$}
	    \State \textcolor{blue}{$\%$ \textit{We assume $R_{D}$ already contains demonstration data}}
	       \State  Sample a mini batch from $R_D$
	        \State Calculate loss with target network and update $\pi$
	        \If{$t\ mod\ p = 0$} $\theta ^{\pi^{'}} \gets \theta^{\pi}$ and $\theta ^{Q^{'}} \gets \theta^{Q}$\EndIf
	    \EndFor
		\For {$t=1,Num\_episodes\times Num\_steps$}
		\State \textcolor{blue} {$\%$ Initial state $s_0$}
				\State Select action $a_t$, execute in the environment and observe reward $r_{t+1}$ and new state $s_{t+1}$
				\State Sample mini-batches from replay buffer $R_D$ and $R_A$
				\State \textcolor{blue}{$\%$ \textit{Update for both minibatches}}
				\State Update the critic with a gradient descent (standard DDPG loss for critic)
				\State \textcolor{blue}{$\%$ \textit{Check if agent's policy is better than heuristic policy for each sample}}
				\State Update the actor with DDPG mean squared loss {$+$} BC loss
				\State {Update $R_{A}$}
				\If{$t\ mod\ p = 0$} $\theta ^{\pi^{'}} \gets \theta^{\pi}$ and $\theta ^{Q^{'}} \gets \theta^{Q}$\EndIf
		\EndFor
	\end{algorithmic} 
\end{algorithm}

\subsection{Mesh Generation}
At the $i$-th step the algorithm separates $W_i$ from initial shape, $B_{i}$ specifies the boundaries of this part, $\{C_k\}_{k=1}^{n_{C}}$ specifies all the slices to reconstruct the shape. $T_j$ denotes slice $j$ of $W_i$, $W_i = \sum_{j=1}^{n_{T}} T_j$, it is formed by cutting $\{C_k\}_{k=1}^{n_{C}}$ according to the boundary $B_{i}$. Dividing the initial shape into smaller parts where each shape is connected (without discontinuities) and also not branched, allows to use more simplified surfacing methods. More concretely, we form a tile of contours from the initial slices and connect them sequentially from the beginning till the end along an axis. For each of the pair of consecutive planes in this tile we find correspondence between contours. While in the optimal case there is only one contour for each plane, this could not be the same for some models with difficult geometries. For finding the correspondence we take into account the center and size of the contours. The final result is consists of the sum of all reconstructed parts $R = \sum_{j=1}^{n_{W}} W_j$.\\
It should be noted that many other iso-surfacing algorithms \cite{10.1145/37402.37422,liu2008surface, Bermano2011, RINEAU2007100} (e.g., marching cubes, octrees, tetrahedras) could also be applied for the surfacing of the parts. However, most of these approaches require additional refinement, which negatively influences on the speed of the reconstruction, especially for the large number of slices. 

\section{Experiments}
\subsection{Experiment Results}
\begin{figure}
  \centering
  \includegraphics[width=0.9\columnwidth]{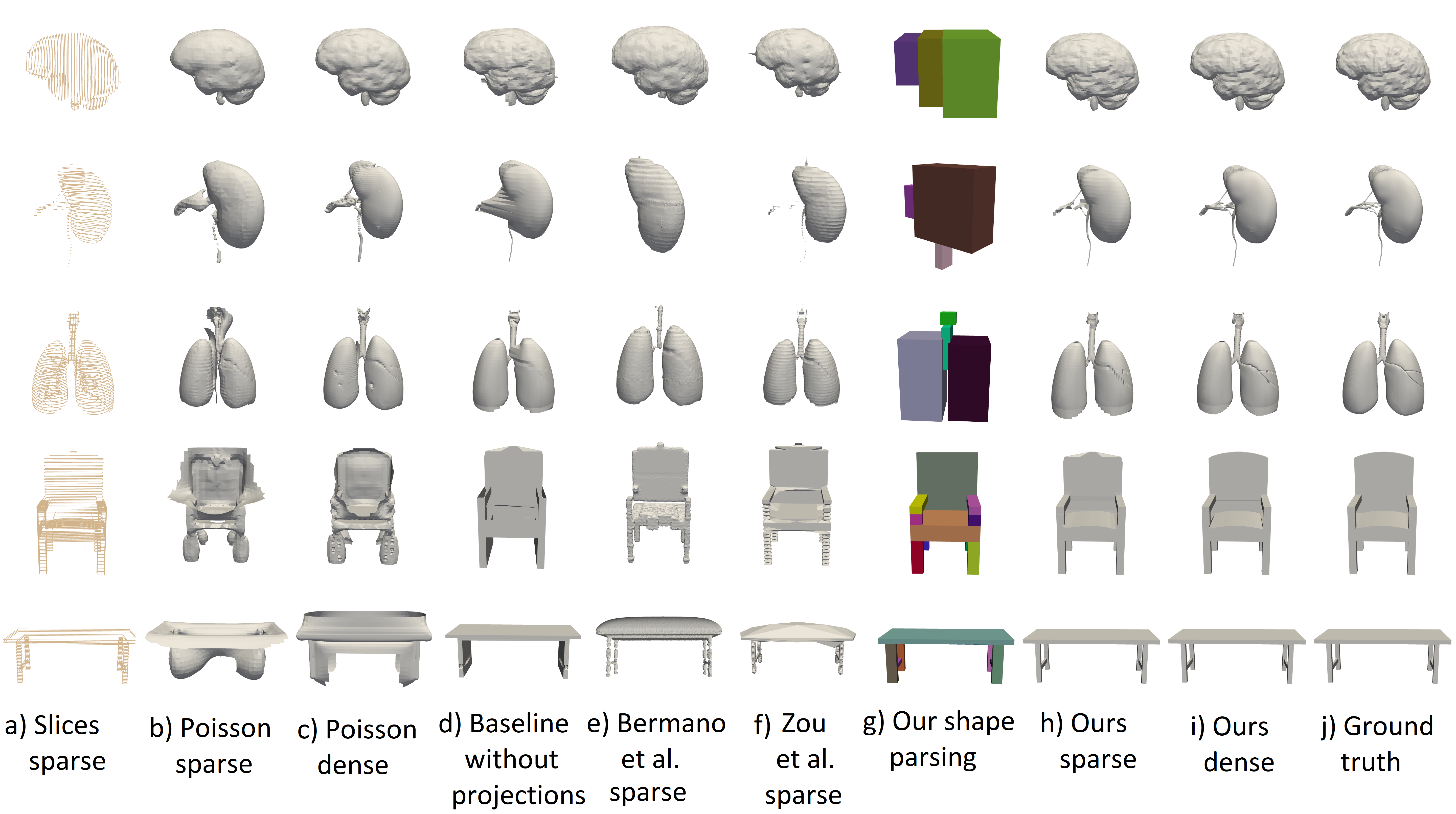}
  \caption{Qualitative comparison of generated results on different initial meshes (brain, kidney, lung, chair, table). The first three models are downloaded from the Internet, while the last two models are taken from ShapeNet dataset. 
  }
  \label{fig:qual_comp}
\end{figure}

\begin{figure}
  \centering
  \includegraphics[width=0.8\columnwidth]{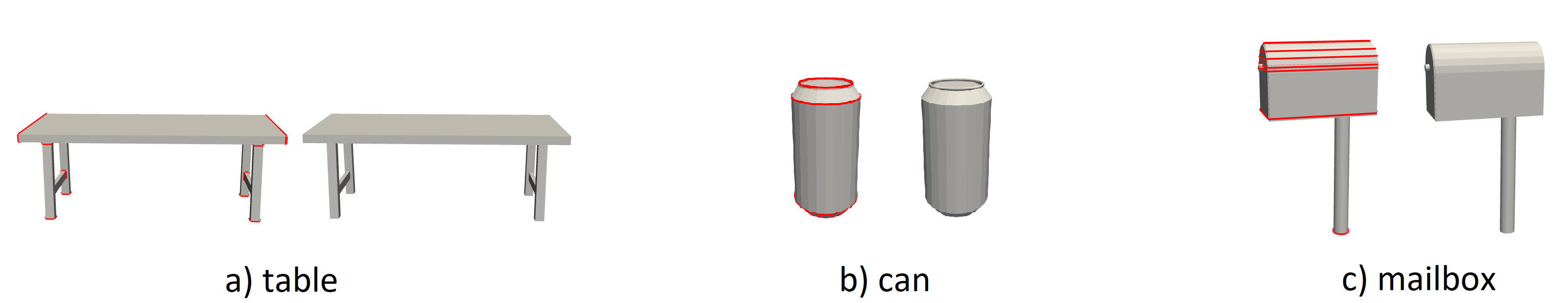}
  \caption{Shape reconstruction from a few slices. Image in the left refers to the reconstructed mesh, and in the right to the ground truth shape for each of three models. The input slices are shown in red.}
  \label{fig:few_slices}
\end{figure}

We train our method on ShapeNet \cite{Chang2015} data set. In order to achieve better generalization we use objects from all classes. In each step the object class and corresponding object from it selected randomly. We follow this strategy for both pre-training and training phases. To assess our agent we conduct two different comparisons with different baselines. In the first experiment we compare our method with the results of slices based surface reconstruction methods (Bermano et al.\cite{Bermano2011} and Zou et al. \cite{Zou2015}), point cloud based method \cite{10.2312:SGP:SGP06:061-070} and baseline method where we use not projection images. Both slices based approaches are build on the level-set extraction of some indicator function. Additionally, Zou et al. \cite{Zou2015} searches for an optimal topology which complies with the user prescribed genus. From this perspective Zou's method is similar to our method. However, our approaches towards solving branching and connectivity problems is completely different. Where we focus on parsing 3D shape into smaller regions which increases reconstruction accuracy. While Bermano's method does not search for the optimal topology it additionally proposes a solution to unknown and multi-labelled regions. Figure \ref{fig:qual_comp} illustrates the result of comparison of these methods on shape reconstruction from the given contours. In this example, 3 out of 5 contours are downloaded from the Internet while 2 others represent ShapeNet objects. Here we consider two different settings of our method for 3D shape reconstruction: sparse input which represent planar cross sections shown in the Figure \ref{fig:qual_comp}, and dense input with two times more slices. As it can be seen from the Figure \ref{fig:qual_comp}, our method captures difficult topology and geometries more accurately for both settings. Also, the quality of reconstruction for all objects are similar for both IoU and CD metrics despite the model was trained only on the ShapeNet data set (Table \ref{table:quan_comp}). It should also be noted that none of these methods generated the results for dense inputs. While our method effectively works with dense and as well as more sparse input. This indicates generalization capability of our method. This should be noted that our IoU calculation takes into account surface correspondence only.\\
As shown in the Figure \ref{fig:qual_comp}, both slices based methods we compare show topological errors despite the input contours are not too sparse. For example, Bermano's method show disconnections almost in all models. Also, there is a problem with handling branching cases in kidney and chair models. Thus requiring more contours to generate thin parts of the shape. Similar disconnection problems are present in the results generated by Zou's method, for chair and table models. While this method handles branching issues in most cases, it adds some additional parts to the shape, in brain and kidney models. We also generate results with point cloud based (Poisson surface reconstruction) method \cite{10.2312:SGP:SGP06:061-070} passing all points in the slices as input. As illustrated in the Figure \ref{fig:qual_comp}, this method generates mediocre outputs for smooth surfaces (e.g. brain, kidney, lung). However, for flat surfaces like chair and table it fails. This is mainly related with the accuracy of point normal estimation for input points. Additionally, Figure \ref{fig:qual_comp} shows results of baseline method without using projections images. However, this method assumes that the input slices are connected. Results generated with this method show some additional parts which not exist in the ground truth mesh. Quantitative results corresponding to this experiment, which includes IoU and Chamfer distance (CD) metrics,  is summarized in Table \ref{table:quan_comp}. Shape reconstruction from a few slices is shown in the Figure \ref{fig:few_slices}. For this type of input slices based surface reconstruction methods, Bermano's method and Zou's method with genus did not output a connected shape. The latter only handles partially when one of the views (as an image) was presented. This also indicates the importance of using orthographic projections.\\

\begin{figure}
  \centering
  \includegraphics[width=0.8\columnwidth]{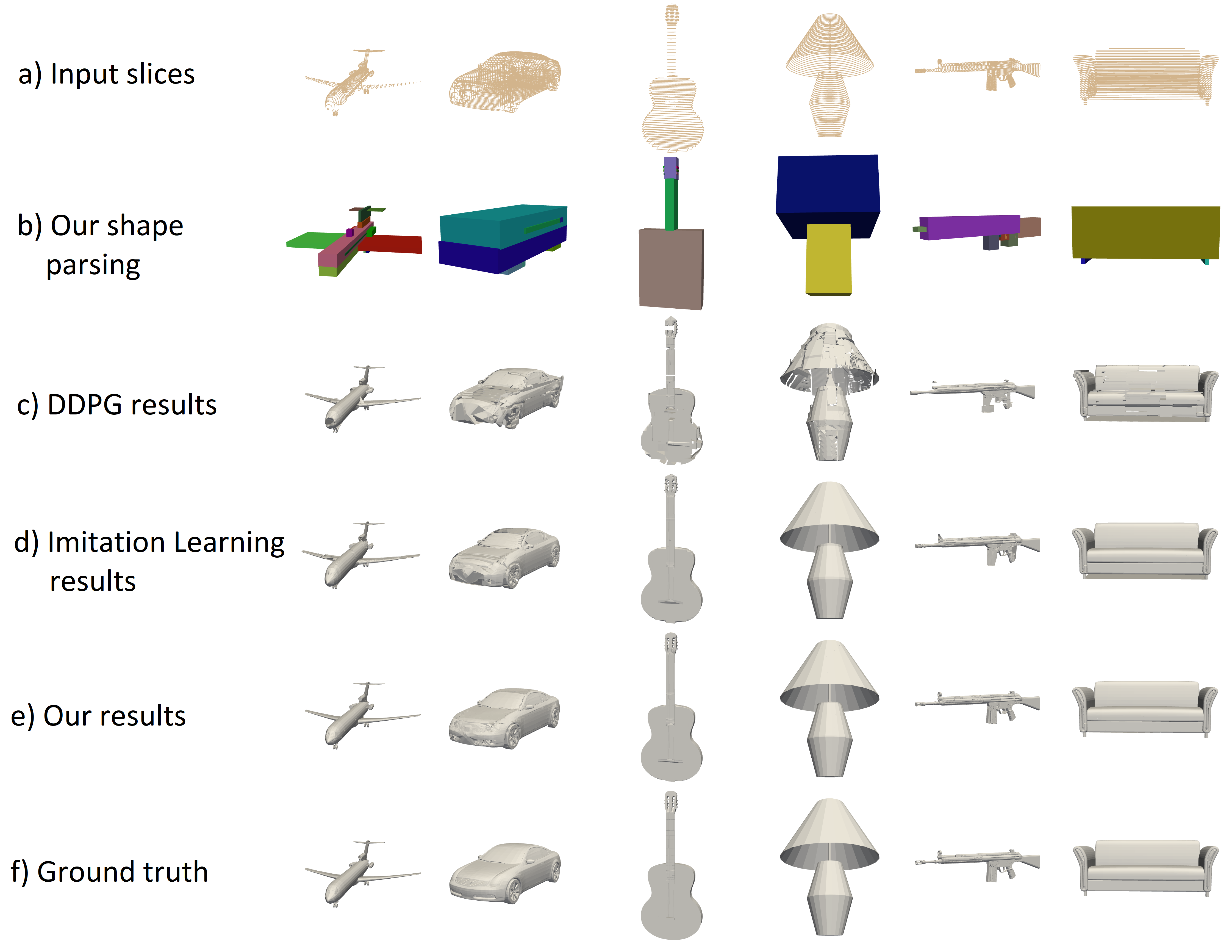}
  \caption{Qualitative comparison of generated results with the ground truth on different initial meshes (airplane, car, guitar, table lamp, rifle, sofa) from ShapeNet data set. Corresponding quantitative results are summarized in Table \ref{table:quan_comp2}.}
  \label{fig:qual_comp2}
\end{figure}

\begin{figure}
\centering
\subfloat[]{{\includegraphics[width=0.26\textwidth]{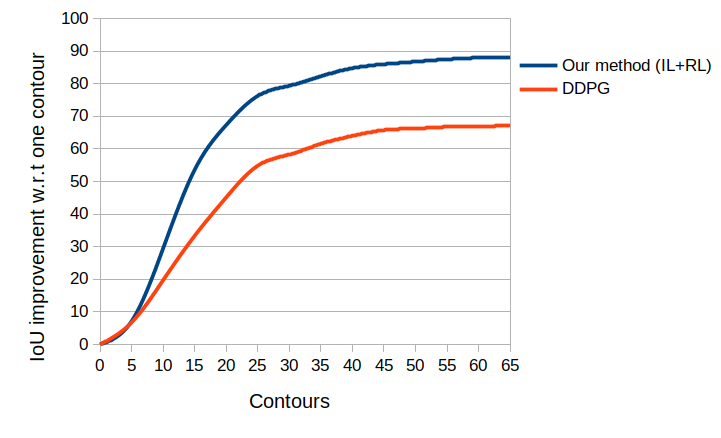} }}%
\qquad
\subfloat[]{{\includegraphics[width=0.26\textwidth]{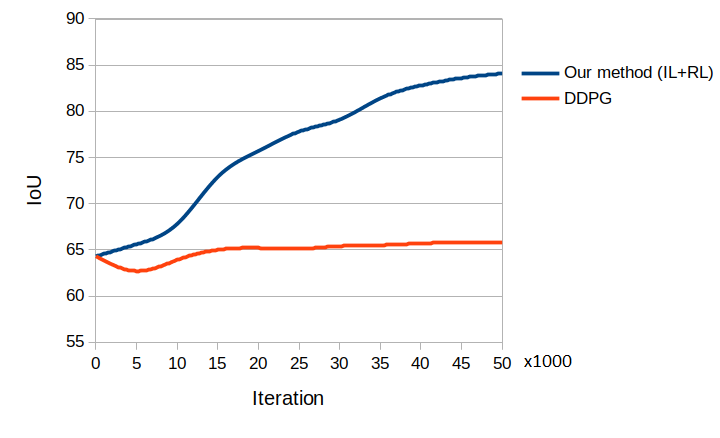} }}%
\qquad
\caption{Comparison of two methods DDPG and Our method (IL+RL) on reconstruction accuracy using IoU metric. (a) IoU improvement w.r.t the number of contours. (b) IoU improvement over iteration.}
\label{fig:DDPGvsOur}
\end{figure}

The second experiment compares our method which includes both IL and RL steps, with IL only and standard DDPG methods on ShapeNet data set object reconstruction. As it shown in the Figure \ref{fig:qual_comp2}, our method effectively captures topological problems compared to the baselines. The standard DDPG method shows the highest reconstruction error (Table \ref{table:quan_comp2}). In most cases, it is due to the reason that it divides the shape into much smaller regions not taking into account the shape topology. Our method which includes only IL step from heuristic policy captures many topological details accurately, especially for simple topologies, e.g. guitar, table lamp and sofa in the example (Figure \ref{fig:qual_comp2}). However, in more complex inputs like airplane, car and rifle it also shows disconnections and branching issues. Our method trained both with IL and RL, handles these problems more accurately for the given input. The quantitative results also reflect this as it is shown in the Table \ref{table:quan_comp2}. We also compare our method (IL+RL) and DDPG on IoU improvement with respect to 1) number of contours and 2) iteration. Figure \ref{fig:DDPGvsOur} illustrates the result of this comparison.
\subsection{Limitation and Future Work}
The current work has a few limitations which are worth addressing. The first noteworthy limitation is that in this paper, we considered to solve the reconstruction problem with only two labels, inside and outside of the surface, in the real world applications there are many examples with multi-labeled objects. This would be an interesting step to extend this work to allow multi-labelled shapes. A second limitation is related with using more complex primitives to describe object boundaries. This would increase the accuracy of the method. Also, we rely on the orthographic projections of the input shape to get information about its topology and geometry. However, for some difficult geometries and topologies this information might not be enough. For such type of input our method might not output very accurate results. Finally, enabling this method to process real data (e.g. from medical device) would be more interesting and more realistic. This can be considered as the future work.   

\begin{table}[]
\centering
\small
\caption{Quantitative comparison of generated results on surface IoU (IoU) and Chamfer-$L_{1}$ (CD) metrics. We measure both metrics with respect to the ground truth mesh. We report numerical results for some of these methods with two different settings: sparse (initial) contours (given in the Figure \ref{fig:qual_comp}a.) and dense contours (with two times more slices) respectively.  
}
\resizebox{\textwidth}{!}{
\begin{tabular}{|p{3.0cm}|p{0.6cm}|p{0.8cm}|p{0.6cm}|p{0.8cm}|p{0.6cm}|p{0.8cm}|p{0.6cm}|p{0.8cm}|p{0.6cm}|p{0.8cm}|}
\hline
\multirow{2}{*}{Method} & \multicolumn{2}{l|}{Brain} & \multicolumn{2}{l|}{\begin{tabular}[c]{@{}l@{}}Kidney\end{tabular}} & \multicolumn{2}{l|}{Lung} & \multicolumn{2}{l|}{Chair} & \multicolumn{2}{l|}{Table}\\
\cline{2-11}
& IoU & CD & IoU & CD & IoU & CD& IoU & CD & IoU & CD\\
\hline\hline
Poisson surface reconstruction \cite{10.2312:SGP:SGP06:061-070} sparse & 0.58 & 0.0177 & 0.13 & 0.0646 & 0.20 & 0.0652 & 0.13 & 0.0891 & 0.07 & 0.1182\\
Poisson surface reconstruction \cite{10.2312:SGP:SGP06:061-070} dense & 0.82 & 0.0106 & 0.65 & 0.0121 & 0.68 & 0.0119 & 0.14 & 0.071 & 0.16 & 0.0695\\
Baseline (no projections) sparse & 0.73 & 0.0125 & 0.57 & 0.0138 & 0.51 & 0.0277 & 0.49 & 0.0259 & 0.64 & 0.022\\
Bermano et. al. \cite{Bermano2011} sparse & 0.29 & 0.0447 & 0.13 & 0.1573 & 0.31 & 0.0493 & 0.16 & 0.0718 & 0.16 & 0.0803\\
Zou et. al. \cite{Zou2015} sparse  & 0.26 & 0.0608 & 0.18 & 0.0972 & 0.27 & 0.0615& 0.54 & 0.0246 & 0.23 & 0.0622\\
Ours sparse  & 0.82 & 0.0106 & 0.78 & 0.0055 & 0.73 & 0.0057 & 0.79 & 0.0118 & 0.83 & 0.0088\\
Ours dense& \textbf{0.88}  & \textbf{0.0074}  &\textbf{0.88} &\textbf{0.0027} &\textbf{0.82} &\textbf{0.0038}& \textbf{0.83} &\textbf{0.0105}& \textbf{0.84} &\textbf{0.0087}\\ 
\hline
\end{tabular}}
\label{table:quan_comp}
\end{table}

\begin{table}[]
\centering
\small
\caption{Quantitative comparison of generated results on ShapeNet data set. We report surface IoU (IoU) and Chamfer-$L_{1}$ (CD) distance metrics. Corresponding graphical comparison is given in Figure \ref{fig:qual_comp2}.  
}
\resizebox{\textwidth}{!}{
\begin{tabular}{|p{2.4cm}|p{1.5cm}|p{1.5cm}|p{1.5cm}|p{1.5cm}|p{1.5cm}|p{1.5cm}|p{1.5cm}|}
\hline
Method-metric & Airplane & Car & Guitar & Table lamp & Rifle & Sofa\\
\hline\hline
DDPG-CD & 0.019 & 0.0228 & 0.0055 & 0.0089 & 0.0066 & 0.0127\\
Ours (IL)-CD & 0.0046 & 0.0135 & 0.0011 & 0.0007 & 0.0028 & 0.0039\\
Ours (IL+RL)-CD & \textbf{0.0018} &\textbf{ 0.0088} &\textbf{ 0.0009} &\textbf{ 0.0007} & \textbf{0.0017} &\textbf{ 0.0012}\\
\hline
DDPG-IoU & 0.68 & 0.61 & 0.76 & 0.66 & 0.79 & 0.62\\
Ours (IL)-IoU & 0.79 & 0.75 & 0.97 & \textbf{0.99} & 0.9 & 0.85\\
Ours (IL+RL)-IoU & \textbf{0.88} & \textbf{0.81} & \textbf{0.99} &\textbf{0.99} &\textbf{0.98} & \textbf{0.91}\\
\hline
\end{tabular}}
\label{table:quan_comp2}
\end{table}

\section{Conclusion}
In this paper, we explored the problem of 3D shape reconstruction from planar cross-sections. Our method to solve this problem, is based on divide-and-conquer strategy to parse the shape into smaller parts that when individually reconstructed and combined boosts accuracy of the generated results. This helps to handle the topological issues like branching and connectivity. For this purpose, we designed RL agent which benefits from IL and also self exploration. While there exist some limitations, such as related with very difficult topologies, the accuracy of the method highlights the effectiveness of this approach.    

\section{Broader Impact}

Our contribution for 3D shape reconstruction from planar cross-sections helps to better understand 3D models from limited data. The impact of this work mainly lie in accurate shape reconstruction from sliced data and 3D shape understanding and analysis. These areas are important research directions which serve to explore and fully understand different properties of 3D objects using computer graphics and vision tools. This work can be applied to solve the real world problems such as shape reconstruction from signals stemming from medical devices. However, the raw data provided from these devices should be converted to a proper input format for this approach. Negative impact could arise when the data not compelling to the input format is used. This model also may result in poor performance when processing input data of the shape with very complex topology and geometry.

{
\small



\bibliographystyle{abbrvnat}
\bibliography{egbib}
}

\clearpage

\appendix



\end{document}